\documentclass[conference]{IEEEtran}
\IEEEoverridecommandlockouts

\usepackage{cite}
\usepackage{amsmath,amssymb,amsfonts}
\usepackage{algorithmic}
\usepackage{graphicx}
\usepackage{textcomp}
\usepackage{xcolor}
\usepackage{mathtools}
\usepackage{amstext} 
\usepackage{array}  
\usepackage{float}
\newcolumntype{L}{>{$}l<{$}} 
\nocite{*}
\def\BibTeX{{\rm B\kern-.05em{\sc i\kern-.025em b}\kern-.08em
    T\kern-.1667em\lower.7ex\hbox{E}\kern-.125emX}}
\begin{document}

\title{Predicting Performance using Approximate State Space Model for Liquid State Machines\\
}

\author{\IEEEauthorblockN{A. Gorad, V. Saraswat, U. Ganguly}
\IEEEauthorblockA{\textit{Indian Institute of Technology, Bombay, India, Phone +91 22 2576 7698}\\
ajinkyagorad@ee.iitb.ac.in}
}
\maketitle

\begin{abstract}
Liquid State Machine (LSM) is a brain-inspired architecture used for solving problems like speech recognition and time series prediction. LSM comprises of a randomly connected recurrent network of spiking neurons. This network propagates the non-linear neuronal and synaptic dynamics. Maass et al. have argued that the non-linear dynamics of LSMs is essential for its performance as a universal computer. Lyapunov exponent ($\mu$), used to characterize the “non-linearity” of the network, correlates well with LSM performance. We propose a complementary approach of approximating the LSM dynamics with a linear state space representation. The spike rates from this model are well correlated to the spike rates from LSM. Such equivalence allows the extraction of a ”memory” metric ($\tau_M$) from the state transition matrix. $\tau_M$ displays high correlation with performance. Further, high $\tau_M$ system require lesser epochs to achieve a given accuracy. Being computationally cheap (1800$\times$ time efficient compared to LSM), the $\tau_M$ metric enables exploration of the vast parameter design space. We observe that the performance correlation of the $\tau_M$ surpasses the Lyapunov exponent ($\mu$),  ($2-4\times$ improvement) in the high-performance regime over multiple datasets. In fact, while $\mu$ increases monotonically with network activity, the performance reaches a maxima at a specific activity described in literature as the “edge of chaos”. On the other hand, $\tau_M$ remains correlated with LSM performance even as $\mu$ increases monotonically. Hence, $\tau_M$ captures the useful memory of network activity that enables LSM performance. It also enables rapid design space exploration and fine-tuning of LSM parameters for high performance.

\end{abstract}


\section{Introduction}

\begin{figure}[ht!] 
\centering
\includegraphics[width=3.5in]{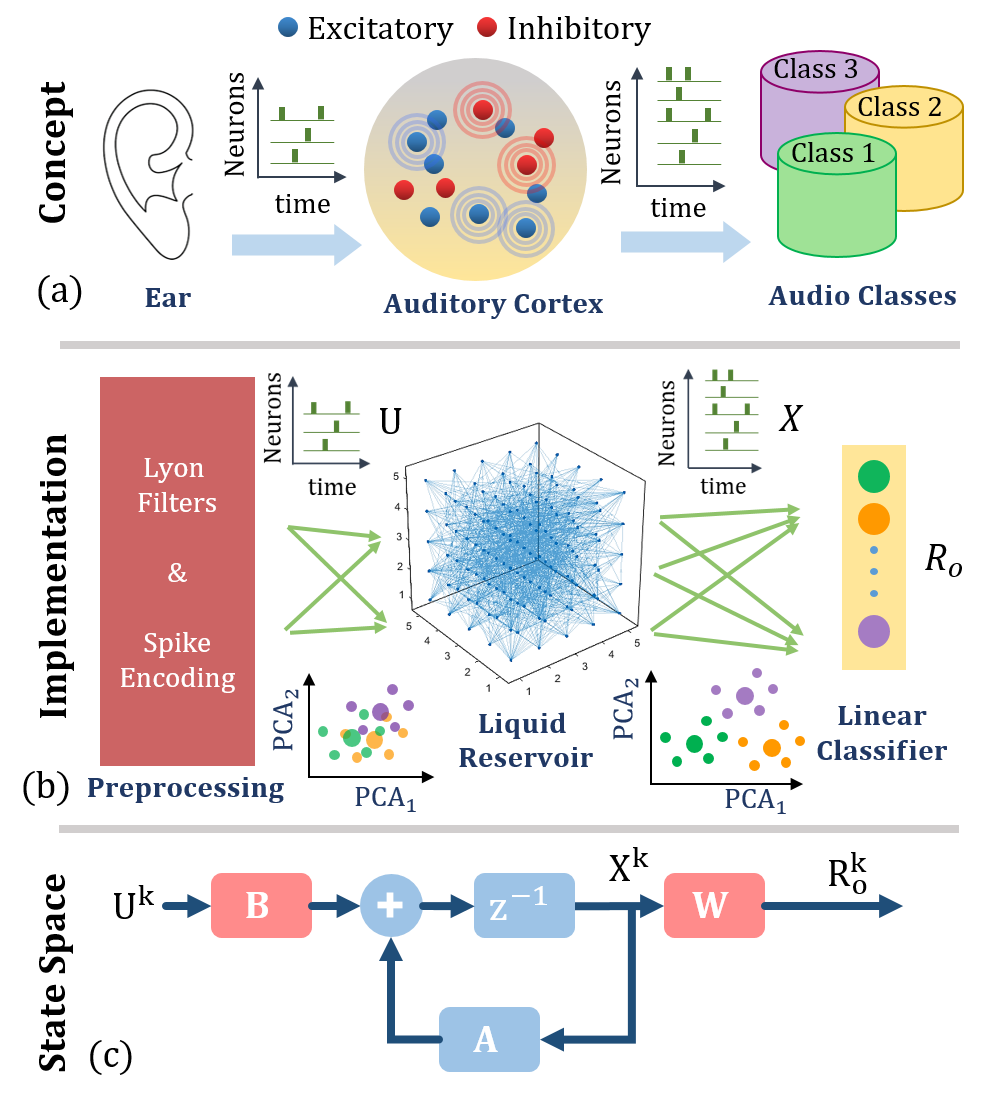}
\caption{Architecture of Liquid State Machine (LSM) showing an example for speech recognition. 
(a) Conceptual description: Spikes generated in the cochlea (ear) travel to the auditory cortex where a randomly connected recurrent neural network acts as a reservoir of a liquid where input spikes produce a "ripple-like" memory in the network. The output of the network transform the non-linearly separable input in low dimensions to a linearly separable output in a higher dimension to enable a simple classifier to perform accurate classification. (b) Implementation of the LSM concept in MATLAB consists of a pre-processing layer (using Lyon's Filter) to generate a spike input, a liquid reservoir of randomly connected recurrent spiking neurons and a linear readout classifier. (c) Discrete Linear State Space approximation of LSM with input mapping \textbf{B}, state transition matrix \textbf{A} and readout weight matrix \textbf{W}.} 
\label{LSM}
\end{figure}

The brain inspired computational framework of Liquid State Machines (LSMs) was introduced by Maass in 2002 \cite{maass2002}. LSM consist of a large recurrent network of randomly connected spiking neurons called the reservoir. The components of this network, namely the neurons and synapses, all follow highly non-linear dynamics. Depending on the extent and strength of connectivity, the reservoir propagates this non-linear dynamics in a recurrent manner. Maass et al. have argued that these non-linear operations performed by LSMs allow it to display high performance and universal computational capabilities \cite{maass2002}. Many applications  are based on the belief that non linearity of the LSMs enable powerful data processing\cite{computationalModel, caseStudy, movementPrediction}.
In \cite{computationalModel} non-linear computations were performed on time series data. In \cite{movementPrediction}, LSM was used for movement prediction task and was shown to perform a non-linear technique of Kernal Principal Component Analysis. Speech recognition, time series prediction, and robot control are few of the many versatile applications for which LSM has demonstrated excellent performance\cite{robot,caseStudy,timeseries,heart}.

LSMs have a vast set of design parameters with no direct relation with the performance. It is also known that, depending on the reservoir connectivity, an LSM may function in the region of low propagation i.e. input spikes may cause a chain reaction in neurons in the LSM to propagate spikes  in the network for a short time. Higher propagation may ultimately lead to chaotic dynamics\cite{metric1}. However, in order to achieve the best performance, the network has to function at an optimal level of activity or, as the literature cites it, at the ``edge of chaos"\cite{edgeOfChaos2}. Many attempts have been made to define metrics to capture the network dynamics that correlates with this trend in performance of LSMs \cite{metric1,lyapunov,separation,separation2,kernel}. Inspired by the idea to capture the extent of non-linear dynamics in the network, Lyapunov exponent is the most successful metric which is well-correlated with the LSM's performance \cite{lyapunov}. Simply put, Lyapunov exponent characterizes the network by measuring the difference in output for two almost identical inputs as the measure of the network's ability to resolve small differences in inputs. As such, it is not an equivalent model for LSM.

State space is an established mathematical modelling framework  especially for time-evolution in linear systems. The framework has a rich legacy of intuition and rigorous techniques for  stability analysis, feedback design for controllers - the many utilities that state space modelling has to offer\cite{NormanNise}. 

In this work, we model the LSM as a linear state space model. We demonstrate that the spike rates obtained from the linear state space representation are well correlated with the spike rates from the exact simulations of LSM. The level of similarity or correlation of the linear model with the exact dynamics depends on the region of performance the LSM is operating in. An LSM operating in the region of high accuracy is indeed well modelled by the linear state space. We show that key advantage of a first order model is that it allows us to evaluate a ``memory" metric $\tau_M$ for the system which shows high correlation with the network performance. $\tau_M$ provides a computationally cheap approach to exploring the design space of the LSMs which have a vast number of design parameters. In addition, we find that the new proposed metric $\tau_M$ surpasses the performance prediction capabilities of the Lyapunov exponent. $\tau_M$ is able to capture the optimal performance achieved as a function of the network activity believed to be occurring at the ``edge of chaos"\cite{edgeOfChaos2}. For fine tuning performance of LSMs, $\tau_M$ is shown to be better than Lyapunov exponent over multiple datasets. 

The paper is organized as follows: Section \ref{background} highlights how LSMs have been implemented to achieve accuracy comparable to the state-of-the-art for the TI-46 speech dataset. In Section III, the state space model is presented to calculate the linear approximation of the LSM behavior and extract the memory metric. Section IV consists of the results and discussions, where $\tau_M$ is calculated for LSMs exhibiting a range of performances. We show that $\tau_M$ significantly outperforms Lyapunov exponent for high performance LSMs.

\section{Background}\label{background}
Liquid State Machine (LSM) framework \cite{maass2002} is analogous to the brain, where, for e.g.  external sensory stimuli like sound is converted to spikes through cochlea in ear. These spikes go to the auditory cortex which form ripples of network activity to enable the conversion of a linearly inseparable input at lower dimensions to a linearly separable output at higher dimensions which are easy to classify (Fig. \ref{LSM} (a)). The implementation of LSM architecture consists of (i) a pre-processing layer where sound is converted to spike trains. (ii) These spikes are introduced to the randomly connected spiking neural network called liquid or reservoir. Here input spikes produce a wave of spikes in the randomly connected recurrent spiking neural network which is similar to a liquid where raindrops cause ripples to propagate (carrying the memory of the raindrops) and eventually fading away. Thus, a network of fixed synaptic weights and delays in the reservoir spreads the input across time and space (i.e. among neurons). The small number of inputs are translated by the large network on neurons in the LSM to a higher  dimensions. This higher dimensional liquid response improves the performance of a simple classifier layer with a learning rule.

\subsection{Speech preprocessing} 
Speech pre-processing stage for an LSM consists of a human ear-like model, namely Lyon's Auditory Cochlear model \cite{LyonModel}. It consists of a cascade of second order filters to produce a response for each channel. The output of each second order filter is rectified and low-pass filtered to get smooth signal. The shape of the second order filters is determined by the quality factor $Q$ and its step size $\Delta_{step}$ \cite{auditoryToolbox}.  Human ears have a large dynamic range (60-80 dB), thus, an Automatic Gain Control (AGC) stage with particular target amplitude $A_{m}$ and timescale $\tau_{AGC}$ is incorporated in the Lyon's model. Since the samples in TI-46 dataset have the sampling rate of 12.5 KHz, the output of this AGC, called cochleogram was decimated in time by decimation factor $df$ of 12 for simulation purposes. This was done with the help of Auditory Toolbox in MATLAB \cite{auditoryToolbox}. This decimated output is converted into spikes trains using BSA algorithm \cite{BSA}. A second order filter $h_{BSA}$ was used for this encoding scheme with the time constants of $\tau_{b1}$ \& $\tau_{b2}$. Finally, 77 spike trains were generated for a corresponding input speech sample in the preprocessing stage. Relevant parameters for this stage are given in Table \ref{prePARAM}.

\begin{equation}	
\begin{split}
h_{BSA} = (e^{-\frac{t}{\tau_{b1}}}- e^{-\frac{t}{\tau_{b2}}})H(t)
\end{split}
\label{BSAfilter}
\end{equation}

\begin{table}[htbp]
\caption{Preprocessing Parameters}
\begin{center}
\begin{tabular}{ c c c c }
\hline
Parameter & Value & Parameter & Value\\ \hline \hline
$Q$ & 8 & $df$ & 12 \\ \hline
$\Delta_{step}$ & 0.25 & $\tau_{b1}$, $\tau_{b2}$  & 4, 1 ms  \\ \hline
$A_{m}$ & 0.0004 & $\tau_{AGC}$ & 32 ms  \\ \hline
\end{tabular}
\label{prePARAM}
\end{center}
\end{table}

\subsection{Liquid Reservoir}
Liquid Reservoir is taken to be a 5x5x5 3D grid of Leaky Integrate and Fire (LIF) neurons with  fraction $f_+$ as excitatory and the rest as inhibitory neurons in the simulations. Neural dynamics are given by \eqref{LIFmodelEqn}-\eqref{LIFmodelEqn2}, where $V$ is the membrane potentia with time constant $\tau_{Neu}$, refractory period $T_{rp}$, threshold voltage $V_{th}$ and elicited spike time $t_s$. $v$ is the synaptic input to the neuron.  $w_{ij}$ is the connection strength from $N_i$ to $N_j$.
\begin{IEEEeqnarray}{CCl}
\frac{\partial V}{\partial t} = -\frac{V}{\tau_{Neu}}+\sum\limits_{i}\sum\limits_{j} w_{ij} v_{j} \label{LIFmodelEqn}\\
V>V_{th}\rightarrow V=0 ~ \forall~ t_s< t<t_s+T_{rp}
\label{LIFmodelEqn2}
\end{IEEEeqnarray}

Neurons are connected with synapses of specific weights by the probabilistic rule given in (\ref{GaussianConnectivity}), where $D(N_1,N_2)$ is the distance between neurons $N_1$ and $N_2$ in the reservoir, $\lambda$ is the effective synaptic distance. $K$ is the connection probability, and it can take values as $K_{EE}$, $K_{EI}$, $K_{IE}$, $K_{II}$. For e.g. $K_{EI}$ corresponds to the probability of connection from excitatory (E) to inhibitory (I) neuron along with the radial dependence \eqref{GaussianConnectivity}.

\begin{equation}
P(N_1,N_2) = K\cdot e^{-\frac{D^2(N_1,N_2)}{\lambda ^2}}
\label{GaussianConnectivity}
\end{equation}

Similarly, synaptic weights $w_{ij}$ can take values $W_{EE}$, $W_{EI}$, $W_{IE}$, $W_{II}$ in the liquid (where subscripts denote E: Excitatory, and I: Inhibitory e.g. $W_{IE}$  denotes weight of the inhibitory to excitatory connection) and are constant in time. Synaptic delays are fixed to $d_{ij}$ for every synapse. Both the excitatory and inhibitory synapses are second order and follow the timescales of ($\tau_{1E}$, $\tau_{2E}$) for excitatory and  ($\tau_{1I}$, $\tau_{2I}$) for inhibitory connections in the reservoir. These timescales correspond to $\tau_1$ and $\tau_2$ for synaptic dynamics given by \eqref{SynModelOrder2}, where $H$ is the unit step function.

\begin{IEEEeqnarray}{CCl}
v = \frac{1}{\tau_1-\tau_2}(e^{-\frac{t-t_{s}}{\tau_1}}- e^{-\frac{t-t_{s}}{\tau_2}})H(t-t_{s})\label{SynModelOrder2}
\end{IEEEeqnarray}

Each input spike train from the preprocessing stage is given to randomly chosen $F_{in}$ neurons in the reservoir with uniformly distributed synaptic weights of $\pm W_{in}$. For these synapses, only excitatory timescales were used. Default network parameters for simulations on TI-46 dataset are given in Table. \ref{ResPARAM}, which are also mentioned in \cite{zhang}.

\begin{table}
\caption{Default Network Parameters}
\centering
\begin{tabular}{c c c c c c }
\hline
Parameter & Value & Parameter & Value  & Parameter & Value  \\\hline \hline
$W_{EE}$ & 3 &  $K_{EE}$ & 0.45 &$\tau_{1E}$ &  8 ms \\\hline
$W_{EI}$ & 6 &  $K_{EI}$ & 0.3  &$\tau_{2E}$ &  4 ms  \\\hline
$W_{IE}$ & -2 & $K_{IE}$ & 0.6  &$\tau_{1I}$ &  4 ms  \\\hline
$W_{II}$ & -2 & $K_{II}$ & 0.15 &$\tau_{2I}$ &  2 ms  \\\hline
$W_{in}$ & $\pm 8$ &$f_+$ & 0.85 & $\lambda$ & 2 \\ \hline
 $F_{in}$&4 & $d_{ij}$ & 1 ms & & \\ \hline
\end{tabular}
\label{ResPARAM}
\end{table}

\subsection{Classifier}
To recognize the class of the input from the multidimensional liquid response, generally, a fully connected  layer of spiking readout neurons is used. Excitatory timescales were used for both excitatory and inhibitory  synapses connecting the reservoir to the classifier. Since the only weight update in an LSM are to happen here, it needs a learning rule which learns the weights to extract useful features from the reservoir. After training, the class corresponding to the most spiked neuron is regarded as the classification decision of the LSM for the given input. 

We briefly describe the biologically inspired learning rule proposed in \cite{zhang}. It uses calcium concentration, $c$ for depicting the activity of the neuron in the classifier to enable selective weight change during a training phase. The calcium concentration for a neuron is defined by a first order equation \eqref{calciumConc} with timescale $\tau_c$. Steady state concentration $c_s$ is approximately given by \eqref{calciumConcSteady} and is the indicator of spike rate ($f_r$) of the neuron. Supervised learning with large forcing current  $I_{teach}$ is used depending on the activity and desirability (or undesirability) of the neuron. The desired (or undesired) neuron is supplied  $I_{\infty}$ (or $ - I_{\infty}$) current to spike more (or less) for the present training input. For each output neuron, this rule uses a probabilistic weight update according to \eqref{learningP}, \eqref{learningN} for the corresponding synapse whenever pre-neuron in the reservoir spikes. A constant weight update $\Delta w$ is added or subtracted, with probability $p^+$ or $p^-$ respectively, to the present synaptic weight depending on the input class and the limit on the activity (decided by $c_\theta$ \& $\Delta c$). Hence, the weight is increased/decreased if the pre-neuron spikes and the classifier neuron is the desired/undesired neuron \eqref{learningP}, \eqref{learningN}. This weight is artificially limited by value $W_{lim}$. The default parameters are mentioned in Table. \ref{NeuPARAM}.

\begin{IEEEeqnarray}{CCl}
\frac{\partial c}{\partial t} = -\frac{c}{\tau_c}+\delta (t-t_s) \label{calciumConc}\\
c_s = \frac{1}{e^{\frac{T_s}{\tau_c}}-1} \approx \tau_c f_{r} \label{calciumConcSteady}\\
 I_{teach} =
  \begin{cases}
             +I_{\infty}\cdot H((c_{\theta}+\delta c)-c), \quad \textit{if desired } \\
            -I_{\infty} \cdot H(c-(c_{\theta}-\delta c)), \quad \textit{if undesired}
  \end{cases}
\label{currentTeach}\\
w \xleftarrow[]{p^+} w+\Delta w ;~ c_{\theta}<c<c_{\theta}+\Delta c \label{learningP} \\
w \xleftarrow[]{p^-} w-\Delta w ;~ c_{\theta}-\Delta c<c<c_{\theta} \label{learningN} 
\end{IEEEeqnarray}

\begin{table}[htbp]
\caption{Neuron \& Classifier Parameters}
\centering
\begin{tabular}{ c c c c c c }
\hline
Parameter & Value & Parameter & Value & Parameter & Value \\ \hline \hline
$\tau_{Neu}$*    & 64 ms & $c_\theta$ & 10 & $W_{lim}$& 8 \\ \hline $T_{rp}$ & 3 ms & $\Delta c$ & 2  & $p^\pm$& 0.1 \\ \hline 
$V_{th}$ & 20 mV & $\delta c$ & 1   &$\Delta w$ & 0.01  \\ \hline
 $I_\infty$  & 10000 &$\tau_{C}$ & 64 ms & &\\
\hline 
\end{tabular}
\label{NeuPARAM}
\end{table}

\subsection{Performance}

We replicated the setup from \cite{zhang}, and matched the state-of-the-art performance for the chosen reservoir size of 5x5x5 for TI46 speech dataset (Table. \ref{perf}).  System is trained and for 200 epochs on 500 TI-46 spoken digit samples. Performance is evaluated using 5 fold testing where accuracy is averaged over the last 20 epochs \cite{zhang}. Samples used in training and testing consisted of a uniform distribution of 50 samples for each digit `0-9' and 100 samples for each speaker, among 5 female speakers.  Time step of 1ms was used in all the simulations.  Simulation was done in Matlab and took a wall-clock time of 14 hours for each run on a Intel Xeon processor running at  2.4 GHz.

\begin{table}[htbp]
\caption{LSM Performance on spoken digit recognition}
\centering
\begin{tabular}{c c c }
\hline
\textbf{Work} &  \textbf{Dataset} & \textbf{Accuracy (\%)} \\ \hline \hline
\textbf{Our} & \textbf{TI-46} & \textbf{99.09} \\\hline
Zhang et al.\cite{zhang} & TI-46 & 99.10 \\\hline
Verstraeten et al.\cite{caseStudy} & TI-46 & 99.5 \\\hline
Wade et al.\cite{wade} & TI-46 & 95.25 \\\hline
Dibazar et al. \cite{dibazar} & TIDigits & 85.5  \\\hline
Tavanaei et al. \cite{spikeSignature} &  Aurora & 91\\\hline
\end{tabular}
\label{perf}
\end{table}

 \subsection{Reservoir-less Network}
 We use a feed-forward fully-connected reservoir-less network to benchmark against the LSM. We train the same preprocessed input on the same classifier. The difference between the performance by this method and the performance by the LSM is the gain/loss in the performance by introducing the reservoir. Generally, LSMs will benefit from the higher dimensional mapping and the recurrent dynamics of the network.

\section{Methodology}
This section describes how the  discrete state space approximation was developed for the LSM dynamics described in the background and the resulting memory metric.

\subsection{State Space Approximation}
To study the dynamics of the spiking trajectories of LSM, we consider the spike rate column vectors for input ($U^k$), reservoir ($X^k$) and readout ($R_o^k$), where each row of the vector corresponds to a instantaneous spike rate at time $k$ for a neuron. Spike rate activity  calculated as the average number of spikes in a rectangular window of 50 ms, and is a column vector where each row corresponds to a neuron. 
This spike rate activity spans a trajectory in a multi-dimensional space (each dimension is represents a different neurons) with time \cite{maassReview}.

The input activity $U^k$ gets mapped to the higher dimensional space of the reservoir. The future activity or state of the liquid $X^{k+1}$ can be written as a function $f$ of its present input $U^k$ and the current state $X^{k}$ \eqref{ResAcc}. Also, since the readout function does not possess any memory, it can be simply the function $f_w$ of reservoir activity $X^k$ \eqref{RoAcc}. 
\begin{IEEEeqnarray}{CCl}
X^{k+1} = f(X^{k},U^{k}) \label{ResAcc}\\
R_o^k = f_w(X^k)        \label{RoAcc}
\end{IEEEeqnarray}

 Dynamics of the LSM are approximated using a state space model which is first order and linear (Fig. \ref{LSM} (c)). Reservoir \eqref{ResAcc} and the readout \eqref{RoAcc} are approximated by \eqref{ResApprox} and \eqref{RoApprox} respectively using constant matrices $A$, $B$ and $W$.

\begin{IEEEeqnarray}{CCl}
X^{k+1} = A \cdot X^{k} + B\cdot U^k \label{ResApprox}\\
R_o^k = W \cdot X^k \label{RoApprox}
\end{IEEEeqnarray}

Let $U$, $X$ and $R_o$  be the actual LSM simulated spike rate over time of all the 10 samples after the training. We denote shifted version of matrix $X$ by 1 time step in future as $X_{+1}$. We estimate $A$,$B$ and $W$ by taking the Moore-Penrose inverse (\textit{pinv}) of the combined matrix of $X$ and $U$ of the system by knowing the spike rate of input, reservoir and readout neurons on chosen 10 samples \eqref{ResPinv}, \eqref{RoPinv}. For a system with M input, N reservoir and L readout neurons, we get size of $A$ to be $N \times N$, $B$ to be $N \times M$ and  $W$ to be $L \times N$. Concatenation of matrices $X$ and $U$ is represented as $[X|U]$.

\begin{IEEEeqnarray}{CCl}
X_{+1} = [A|B]\cdot[X|U]^\intercal \\
\Rightarrow [A|B] = X_{+1} \cdot pinv([X|U]^\intercal) \label{ResPinv} \\
R_o = W \cdot X
\Rightarrow [A|B] = R_o \cdot pinv(X) \label{RoPinv}
\end{IEEEeqnarray}

Once $A$, $B$ and $W$ are determined, Knowing only the input $U$, we estimate the $\hat{X}$ and, finally, $\hat{R_o}$ from the estimated $\hat{X}$. To evaluate the effectiveness of state space modelling of an LSM, we find the correlation coefficient of the actual response $X$ with predicted response $\hat{X}$ knowing the input ($U\rightarrow \hat X$). We also evaluate other combinations of prediction to see how well this approximation holds. Forward combinations include $U\rightarrow \hat X$, $X\rightarrow \hat R_o$ and $U\rightarrow \hat X \rightarrow \hat R_o$. Reverse combinations include $R_o \rightarrow \hat X$, $X\rightarrow \hat U$ and $R_o \rightarrow \hat X \rightarrow \hat U$. This correlation coefficient qualifies the ability of state space to model LSM  (discussed in Section IV). 

\subsection{ Concept of memory}
For an $N$ dimensional state space represented by $X$ given by \eqref{NDss} having N time constants $\tau_i$. Memory of such a system can be defined as the mean of the time constants $\tau_i$ \eqref{NDmem}.

\begin{IEEEeqnarray}{CCl}
\dot X = - diag(\frac{1}{\tau_1},\frac{1}{\tau_2}, ...,\frac{1}{\tau_N})\cdot X \label{NDss}\\
\tau_M = \frac{1}{N}\sum_{i=1}^{N} \tau_i \label{NDmem}
\end{IEEEeqnarray}

A discrete first order system with time constant $\tau_i$ is represented as \eqref{eq1D} using Euler method. Here, $h$ is the time step of the discretized system.
    
\begin{IEEEeqnarray}{CCl}
x_i^{k+1} = (1-\frac{h}{\tau_i}) x_i^{k} \label{eq1D} 
\end{IEEEeqnarray}

For a system matrix A of size $N\times N$ in a discrete state space system which defines the time dynamics, we get its diagonal entries in vector $a$. From this, we propose the memory metric $\tau_M$  for an approximate model of the reservoir \eqref{ResApprox} to be \eqref{memoryMetric}. In our case, discrete time step h is 1ms.
\begin{IEEEeqnarray}{CCl}
a = diag(A)\\
\tau_M = \frac{1}{N}\sum_{i=1}^{N} \frac{h}{1-|a_i|}
\label{memoryMetric}
\end{IEEEeqnarray}

This memory metric is calculated and  and its relation with performance is explored (presented in Section IV). 
It is also compared to the previously identified state of the art performance prediction metric Lyapunov exponent ($\mu$) \cite{lyapunov}. Intuitively, Lyapunov exponent simply characterizes the chaotic behaviour of the network by measuring the difference in output for two almost identical inputs as the measure of the network's ability to resolve small differences in inputs. As such, it is not an equivalent model for LSM. It is calculated as the average over the scaled exponents ($\mu_i$) for classes $i=1,2,...,10$, where for each class i, $\mu_i$ is calculated from two samples $u_{1i},~u_{2i}$ and their reservoir response $x_{1i},~x_{2i}$ using \eqref{LyapunovCalc}.
\begin{IEEEeqnarray}{CCl}
\mu_i=ln\frac{||x_{1i}(t)-x_{2i}(t)||}{||u_{1i}(t)-u_{2i}(t)|| }
\label{LyapunovCalc}
\end{IEEEeqnarray}

 \subsection{Simulation Methodology}
 The reservoir in an LSM has a large number of parameters defining its design space. Study by \cite{edgeOfChaos} identified few key parameters, which include synaptic scaling $\alpha_w$ and effective connectivity distance $\lambda$.  We vary the given synaptic weights by a constant factor $\alpha_w$. We simulate the LSM for speech recognition task using  TI-46 dataset and evaluate the performance over 12 different $\alpha_w$, each  comprising of 4 randomly generated structures for the same parameter settings.

\section{Results and Discussion}

\begin{figure}[b!] 
\centering
\includegraphics[width=3.5in]{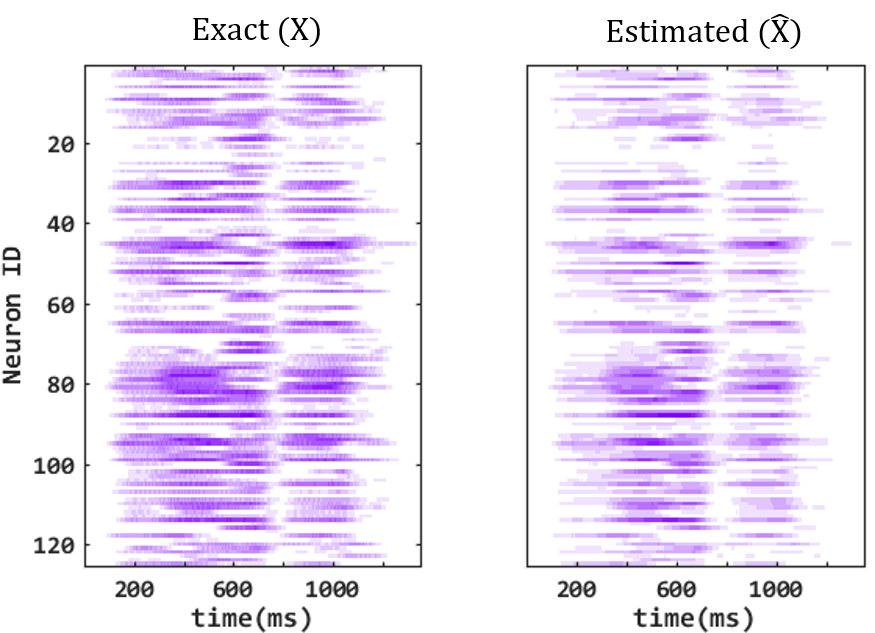}
\caption{Well correlated (Pearson Correlation Coefficient i.e. PCC = 0.92) reservoir spike rates for  LSM $X$ and state space estimation $\hat X$ for each neuron in the reservoir as a function of time for a speech sample from TI-46 dataset.}
\label{estimatedRes}
\end{figure}

\begin{figure}[b!] 
\centering
\includegraphics[width=3.5in]{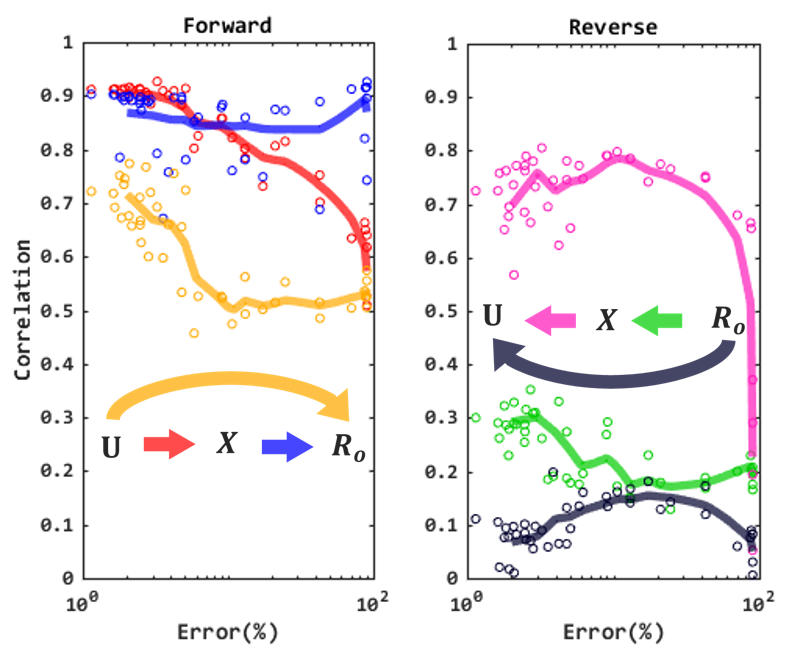}
\caption{Average correlation coefficient between LSM and State Space Model spike rates vs. error (log scale) for all the possible transformations on all 500 samples of TI-46 dataset. Forward correlation are stronger than backward correlations. In the forward transformations, correlations improve in low error (high accuracy) regime.  }
\label{allcorr}
\end{figure}

\begin{figure*}[ht!] 
\centering
\includegraphics[width=7in]{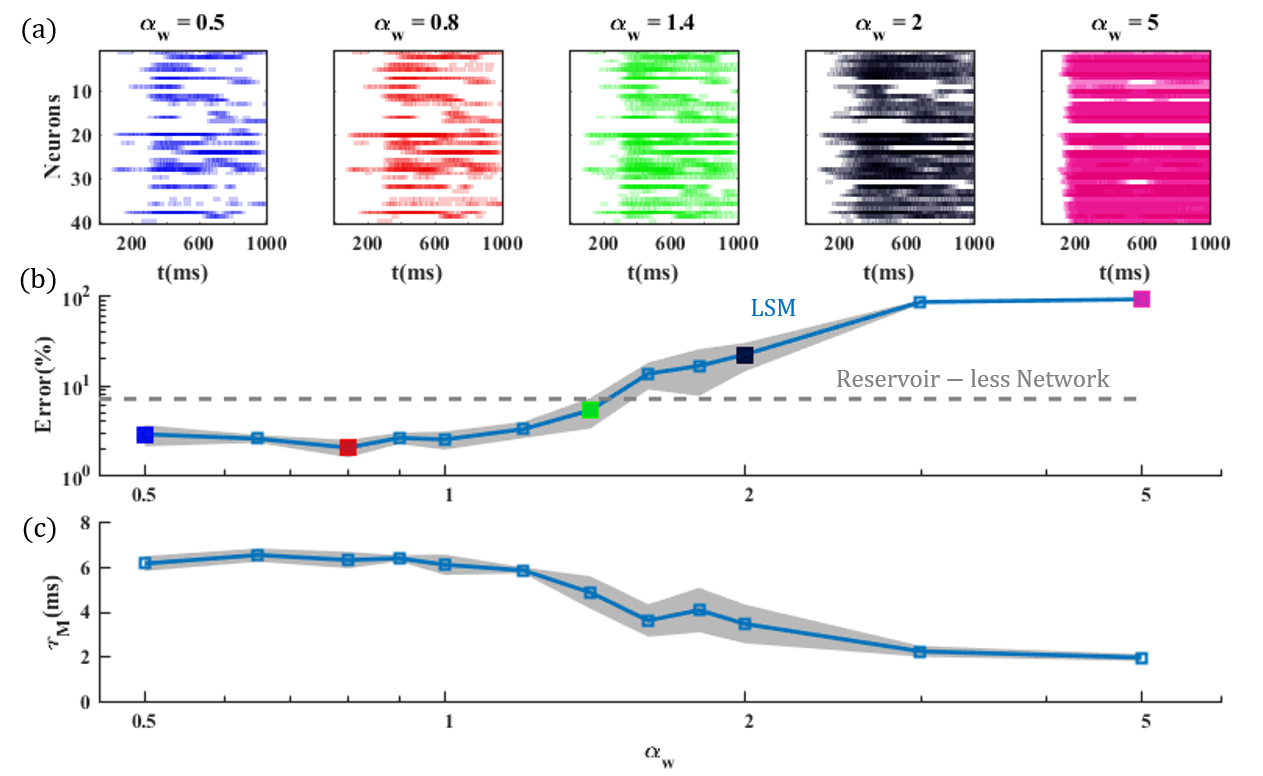}
\caption{(a) Spike rate in the reservoir vs. synaptic scaling ($\alpha_w$) for 40 neurons resulting from a TI-46 speech sample. At the optimal value of $\alpha_w$ = 0.8, the spike propagation is optimal which related to high accuracy. At lower $\alpha_w$ ($<$ 0.8), the spike propagation is weak and reduces accuracy. For higher $\alpha_w$ ($>$ 0.8), the spike propagation is excessive, leading to degraded performance. Higher $\alpha_w$ = 5, leads to relentless spiking.  (b) Error vs. synaptic weight scaling and (c) Memory metric $\tau_M$ vs. synaptic scaling both show non-monotonic but correlated dependence.   One sigma variation over different runs is also plotted in grey in (b) and (c). The performance of a reservoir-less network, where the input is directly connected to the classifier, is shown in comparison to an LSM.}
\label{ActivityMemoryPerfvsaw}
\end{figure*}

\subsection{Similarity to State Space}

We estimate the reservoir spikes $\hat X$  and find them to be well matched with the actual spike rate $X$ for the transformation $U\rightarrow \hat X$ using the state space approximation with Pearson Correlation Coefficient (PCC) of 0.92. Figure \ref{estimatedRes} shows the visual representation of actual reservoir spike rate and the estimated spike rate for each neuron in the reservoir in the high performance region of operation.

\begin{figure}[t!] 
\centering
\includegraphics[width=3.5in]{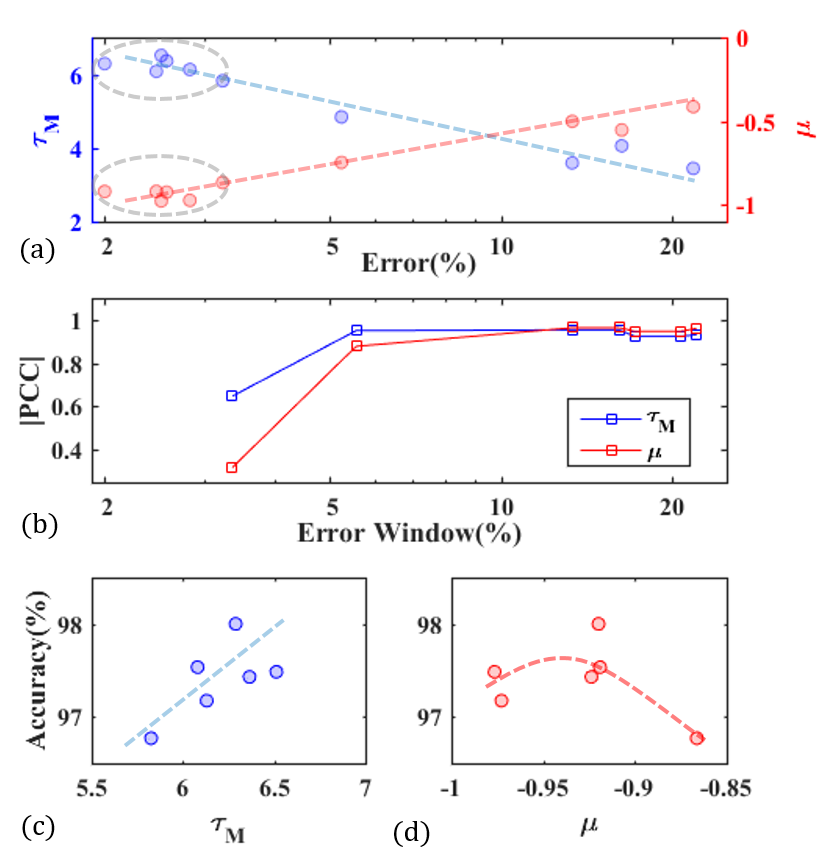}
\caption{(a) Correlation of $\tau_M$ and $\mu$ with error for TI-46 speech dataset. (b) For overall error range, the correlation values are comparable for $\tau_M$ and $\mu$ (0.96). Correlation of $\tau_M$ (0.64) with accuracy is 2$\times$ higher than that of $\mu$ (0.31) at lower errors is also seen by the low error expanded view behaviors in (c) and (d) of the encircled regions in (a)}
\label{tauVsLyapunov}
\end{figure}

Further, correlation coefficients for all model estimated spike rates with exact spike rates are evaluated as a function of performance for both the forward and reverse pathways (Fig. \ref{allcorr}). Forward estimation gave higher correlation values than reverse estimation. 
Further, when estimating the reservoir spikes from the input spikes ($U\rightarrow \hat X$), we find that this model is a good fit when error is small. In other words, the mapping of LSM to state space is more accurate when the output performance of the network is high.

We obtain the state space model estimation of the readout $f_w$ by $W$ with correlation greater than  0.85 for all ranges of error ($X \rightarrow \hat R_o$). The overall transition from $U \rightarrow \hat X \rightarrow \hat R_o$ is the combination of $U\rightarrow \hat X$ \& $X\rightarrow \hat R_o$ and hence, the correlation between $R_o$ and $\hat R_o$ given the input $U$  is lower than both.

The mapping from readout to reservoir ($R_o \rightarrow \hat X$) gives weak correlation coefficients in reverse estimation. This is intuitively expected as the readout need not represent all the information in the liquid mainly because the classifier has a lower dimensionality of 10 compared to 125 in the reservoir. In comparison, when we estimate the input (of dimensionality 77 in these simulations) from the reservoir $X \rightarrow \hat U$, the correlation is close to 0.8, and overall estimation correlation $R_o \rightarrow \hat X \rightarrow \hat U$ is less than both $R_o\rightarrow \hat X$ \& $X\rightarrow \hat U$.

\begin{figure}[hb!] 
\centering
\includegraphics[width=3in]{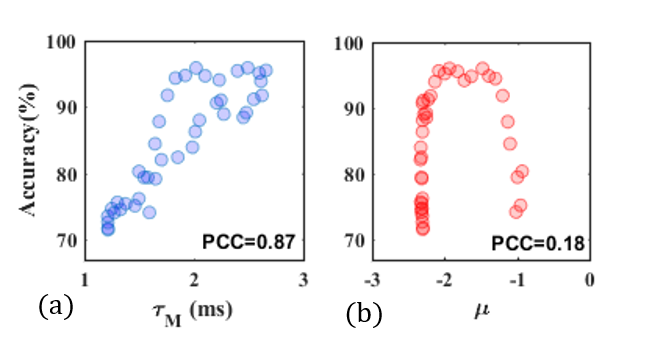}
\caption{Performance dependency with our memory metric $\tau_M$ and Lyapunov exponent $\mu$ for generated poisson spike dataset. Synaptic weight scaling $\alpha_w$ was varied. Values are shown for average accuracy over last 20 epochs from total of 100 epochs for 2-fold testing. Significantly better correlation ($>4\times$) to performance is observed in case of $\tau_M$ (0.87) compared to Lyapunov exponent (0.18) for the network.}
\label{PoissonAcc}
\end{figure}

\subsection{Performance as a function of Network Activity}
We calculate the performance and the memory metric in different regimes of the LSM operation. For different reservoir weight scaling $\alpha_w$ (also called synaptic weight scaling) we obtain low-propagation (sparse activity $\alpha_w=0.5$), significant-propagation (normal activity $\alpha_w=0.8$), in-chaos (high activity $\alpha_w=2$) and saturation (all neurons spiking $\alpha_w=5$), shown in Fig.\ref{ActivityMemoryPerfvsaw} (a). Performance of the LSM for various synaptic weight scaling is found (Fig. \ref{ActivityMemoryPerfvsaw} (b)) and the
memory metric $\tau_M$ of the corresponding state space modelled system is found using \eqref{memoryMetric}, and its variation against synaptic weight scaling is shown in Fig. \ref{ActivityMemoryPerfvsaw} (c).

\begin{figure}[t!] 
\centering
\includegraphics[width=3.5in]{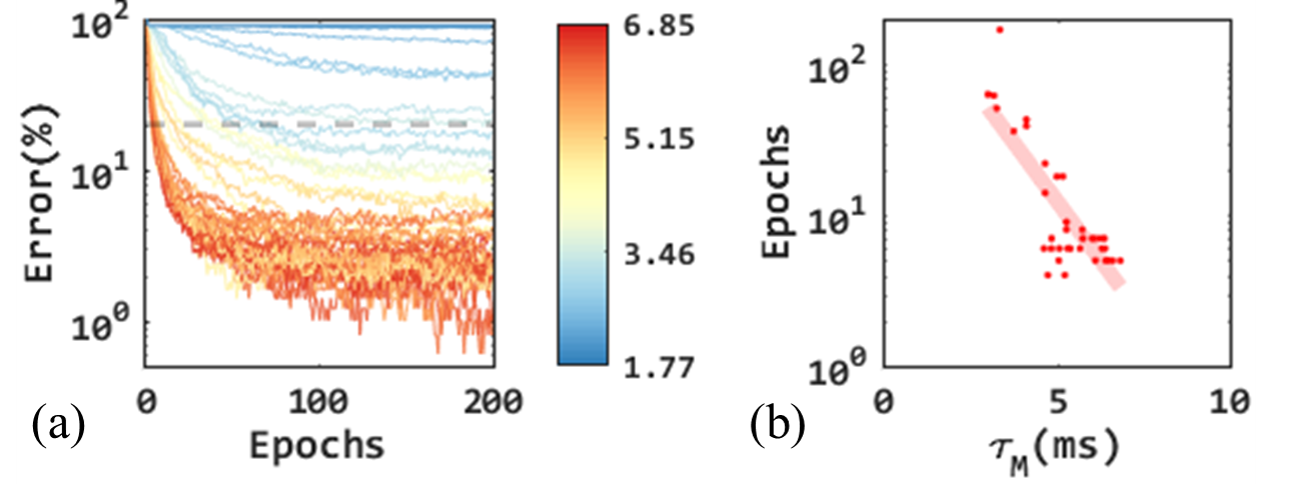}
\caption{(a) Error vs epochs with its memory metric $\tau_M$ (ms) (color) for TI-46 dataset. Dashed grey line indicates error at which points are plotted in (b). (b) Number of epochs required for achieving 80\% accuracy. Lower epochs are required to achieve same accuracy for higher $\tau_M$}
\label{EpochsTau}
\end{figure}

For low-propagation in the reservoir where $\alpha_w$ is 0.5, there is mainly higher dimensional mapping of the input and limited dynamics take place in the reservoir. This higher dimensional mapping gives a boost of 5\% to the accuracy 93\% of reservoir-less approach (Fig. \ref{ActivityMemoryPerfvsaw}). Increasing the weights  ($\alpha_w=0.8$) allows input spikes to propagate through the reservoir and hence allows memory dynamics, which contribute to the performance.
This results in the reduction of error to 1.5\%. The size of the LSM and activity together maximize the accuracy which corresponds to increase in the memory metric.

However, increasing activity does not always correspond to good performance, and hence should also not contribute to the memory of useful information present in the system. Our proposed memory metric decreases because of the increasing activity, as the disorder increases and the LSM enters chaos. In saturation regime ($\alpha_w$ = 5), it is evident that there is not much useful information as the pattern is lost permanently.

We think that the state space approach helps in characterizing the memory of LSM, because,  useful information is related in time by its past activity and the current input. This approximation of trajectories provides an estimate of how the system evolves with the addition of new input. As the LSM enters chaotic regimes, the information is actually destroyed due to disturbance of this trajectory, this is highlighted by the breaks in the trajectory and cannot be captured by the smooth state space transition.

\subsection{Comparison with Lyapunov Exponent }

Following from the above discussion, the key property of the proposed memory ($\tau_M$) metric that emerged on varying the synaptic weight scaling ($\alpha_w$) is that $\tau_M$ has very high correlation with the recognition performance, i.e. large memory results in higher accuracy (Fig. \ref{tauVsLyapunov} (a)). A more significant trend is observed when we focus on the high performance region. The PCC of $\tau_M$ with accuracy was found to be 0.93 over all possible error values and it increased to 0.95 for points in close proximity to small error (Fig. \ref{tauVsLyapunov} (b)). As a comparison benchmark, we also calculated the absolute PCC with accuracy of the previously identified \cite{lyapunov} state-of-the-art performance prediction metric, Lyapunov exponent ($\mu$) as 0.95 which rapidly dropped down as we focused on the high performance region of operation (Fig. \ref{tauVsLyapunov} (b)). In other words, the overall performance correlation for both the metrics is at par, however, $\tau_M$ performs better than $\mu$ for low error regions. This is due to the monotonic nature of Lyapunov exponent with respect to synaptic scaling. In general the accuracy of LSMs increase as the activity increases but then reduces with the onset of chaos. $\tau_M$ captures this optimal activity threshold precisely increasing its utility over the Lyapunov exponent. 

In addition, this behavior of $\tau_M$ is general and can be extended to other datasets. We adopt a well known strategy for generating another test dataset \cite{datasetPoisson,edgeOfChaos}.
We construct 10 input spike train templates of 40 Hz poisson distributed, each comprising 10 spike channels with sample length of 200 ms. From each of them we generate 50 spike patterns, where spike of each template is jittered temporally with standard deviation of 16 ms. Any resulting spikes outside the window of sample are removed.
We train this dataset for 100 epochs on the LSM with a different set of reservoir parameters as mentioned in \cite{edgeOfChaos} and vary the synaptic scaling from 0.1 to 4. With this dataset, we found the PCC of memory metric to the performance to be 0.87 (Fig. \ref{PoissonAcc} (a)), which is significantly greater than PCC of 0.18 (Fig. \ref{PoissonAcc} (b)) for the Lyapunov exponent. This suggests memory metric $\tau_M$ is a better measure for performance with increased generality across datasets.

Another attractive property of $\tau_M$ is the associated time efficiency of the system to achieve a given accuracy. Fig. \ref{EpochsTau} (a) shows a large density of high $\tau_M$ systems achieving greater accuracy more quickly (in lower number of epochs). For performance to fall below a specified error, we get an exponential relationship between the number of epochs required and the associated memory metric of the system (Fig. \ref{EpochsTau} (b)). In other words, the memory metric has a direct impact on the learning rate of the classifier with faster learning being enabled in high $\tau_M$ systems.

\subsection{Computational Efficiency}

One simulation of exact LSM takes 1.5 hrs on average, if the system is trained for only 20 epochs. In comparison, one $\tau_M$ extraction from the approximate state space modelling takes only 3 seconds which is an 1800$\times$ speed up. Further,  $\tau_M$ calculation is 2$\times$ faster than the calculation for the Lyapunov exponent.

\subsection{Design Space Search}

\begin{figure}[ht!] 
\centering
\includegraphics[width=3.5in]{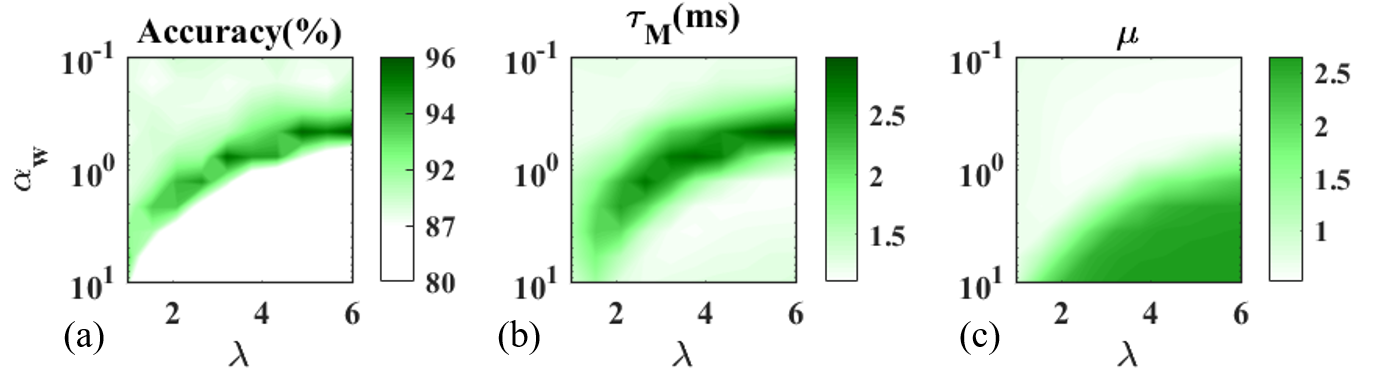}
\caption{The memory metric is used for design space search using Poisson Spikes dataset, where space is explored over reservoir synaptic scaling $\alpha_w$ and effective connectivity $\lambda$. On the grid of 10x10, shown are the (a) LSM simulation result with average accuracy over last 5 epochs from total 20 epochs with 2-fold testing  (b) Our memory metric $\tau_M$ (c) Lyapunov exponent $\mu$. We observe that high performance regions are clearly highlighted by the memory metric, while Lyapunov exponent does not captures the optimal space of $\alpha_w$ and $\lambda$ as it shows a monotonic trend. Higher PCC (5x) with performance was observed for $\tau_M$ (0.90) compared to $\mu$ (0.18).}
\label{SpaceSearch}
\end{figure}

Given the performance predictive properties of $\tau_M$, the design space exploration is greatly simplified for LSMs. As highlighted in the simulation methodology, LSMs can be tuned by varying a large number of parameters. To highlight the utility of $\tau_M$ in this regard, we use previously identified synaptic scaling ($\alpha_w$) and effective connectivity $\lambda$ as the key performance parameters \cite{edgeOfChaos} to explore the accuracy achieved. We compare the accuracy with the calculated $\tau_M$ and $\mu$ over the same parameter space. The accuracy shows a region of maxima (dark green - Fig. \ref{SpaceSearch} (a)) and falls off on either side over the parameter space. This behavior can also be seen for the $\tau_M$ obtained over the same parameter space (Fig. \ref{SpaceSearch} (b)). Hence, calculation of $\tau_M$ from state space approximation provides an efficient method to identify the correct parameter values for high performing LSM. Again, as discussed earlier in Fig. \ref{tauVsLyapunov}, the Lyapunov exponent does not capture the performance optima unlike $\tau_M$ and is monotonic in nature with network activity. Correlation was found to be 0.90 for the memory metric and 0.18 for the Lyapunov exponent in the high performance region (Accuracy$>85\%$) for the design space search conducted over $\alpha_w$ and $\lambda$ (Fig. \ref{SpaceSearch}).

\section{Conclusion \& Future Work}
It is widely believed that recurrent network of non-linear elements producing highly non-linear information processing to enable high performance in recognition tasks. In this paper, we present an alternative interpretation of LSM where we approximate it with a linear state space model - a well-established mathematical framework. We demonstrate high correlation of the model response with the LSM. Equivalence with state space allows the definition of a memory metric which accounts for the additional performance enhancement in LSMs over and above the higher dimensional mapping. The utility of $\tau_M$ as a performance predictor is general across datasets and is also responsible for time efficient learning capabilities of the classifier at the output. We compare and highlight the advantages of $\tau_M$ over the existing state-of-the-art performance predictor metric Lyapunov exponent $\mu$. where a 2-4$\times$ improved correlation of accuracy is observed for $\tau_M$.  $\tau_M$ captures the maximum performance for optimum parameters in the design space where the network has optimal activity is at the ``edge of the chaos''. In contrast, Lyapunov exponent is monotonic with network activity - resulting in poor performance prediction at high performance. Further, the computational efficiency of the state space model (1800$\times$ compared to LSM) to compute $\tau_M$ enables rapid design space exploration and parameter tuning.

\end{document}